\documentclass[letterpaper, 10 pt, conference]{ieeeconf}
\IEEEoverridecommandlockouts
\usepackage{graphicx}
\usepackage{amsmath}
\usepackage{amssymb}
\usepackage{url}
\usepackage{hyperref}
\hypersetup{
    colorlinks=true,
    linkcolor=black,
    citecolor=black,
    filecolor=black,
}
\usepackage[all]{hypcap} 

\title{\LARGE \bf
2D and 3D Grasp Planners

for the GET Asymmetrical Gripper
}

\author{Andrew Goldberg$^1$, Ethan Ransing$^1$, Anton Kourakin$^1$, Cael Magner$^1$, Edward H. Adelson$^2$, Ken Goldberg$^1$
\vspace{0.1cm}\\
\url{https://apgoldberg1.github.io/GET_Planning}
\vspace{-0.2cm}
\thanks{$^1$The AUTOLab at UC Berkeley, $^2$Massachusetts Institute of Technology}  
\thanks{Emails: \texttt{\{apgoldberg, ethan.ransing, akourakin, cael, goldberg\}@berkeley.edu},  
\texttt{\{adelson\}@csail.mit.edu}}  
\thanks{AUTOLab Website: \url{https://autolab.berkeley.edu/}}
}

\begin{document}

\maketitle
\thispagestyle{empty}
\pagestyle{empty}

\begin{abstract}
In this paper, we introduce GET-2D-1.0, a fast grasp planner for the GET asymmetrical gripper that operates from a single-view RGB-D image, using the Ferrari-Canny metric and a novel sampling strategy, and GET-3D-1.0, a mesh-based method using a 3D gripper model and ray-tracing. 
We evaluate both grasp planners against baselines with physical experiments, which suggest that GET-2D-1.0 can improve over a bounding box baseline by over 40\% in lift success, shake survival, and force resistance. Experiments with GET-3D-1.0 suggest slight improvement compared to GET-2D-1.0 on lift success and shake survival, but are more computationally expensive, averaging 17 seconds of planning compared to 683 ms for GET-2D-1.0.
\end{abstract}

\section{INTRODUCTION}

Traditionally, industrial robots have relied on symmetric parallel-jaw grippers due to their simplicity in actuation and geometric modeling~\cite{cut1985}. 
However, with standard narrow jaws, the two antipodal point contacts can act as a pivot axis with limited resistance to applied torques.

When a robot arm attempts to manipulate a tool, lift an off-center payload, or engage in tasks like sweeping or drawing, the lack of torque resistance can result in objects slipping. 
When the instability of narrow jaws is addressed by adopting symmetric wide jaws, they will have a higher probability of unintentional collisions. 
Beyond reachability, overhead or wrist-mounted vision systems rely on a clear line of sight to verify object pose, and wide jaws inherently occlude the field of view. 

In 2025, Burgess and Adelson introduced the Grasp EveryThing (GET) gripper, a 1-DOF, asymmetric design whose V-shaped jaw geometry enables three points of contact that can form a stabilizing lever arm against torque disturbances~\cite{burgess2025get}. Burgess and Adelson developed the asymmetrical GET gripper and explored its usefulness using teleoperation, i.e., with a human operator. Grasp planners designed for standard parallel-jaw grippers, vacuum-based suction cups, or multi-finger hands~\cite{mah2017, mah2018, sal1982} do not exploit the unique geometry of the GET gripper. 
In this paper, we develop and implement two algorithmic grasp planners and evaluate them using a robot arm and digital force sensor.

This paper makes the following contributions:

1) GET-2D-1.0: A fast 2D grasp planner for the GET gripper that operates from a single-view RGB-D image, using force closure, the Ferrari-Canny metric, and a novel grasp sampling strategy.

\begin{figure}
    \centering
    \includegraphics[width=.96\linewidth]{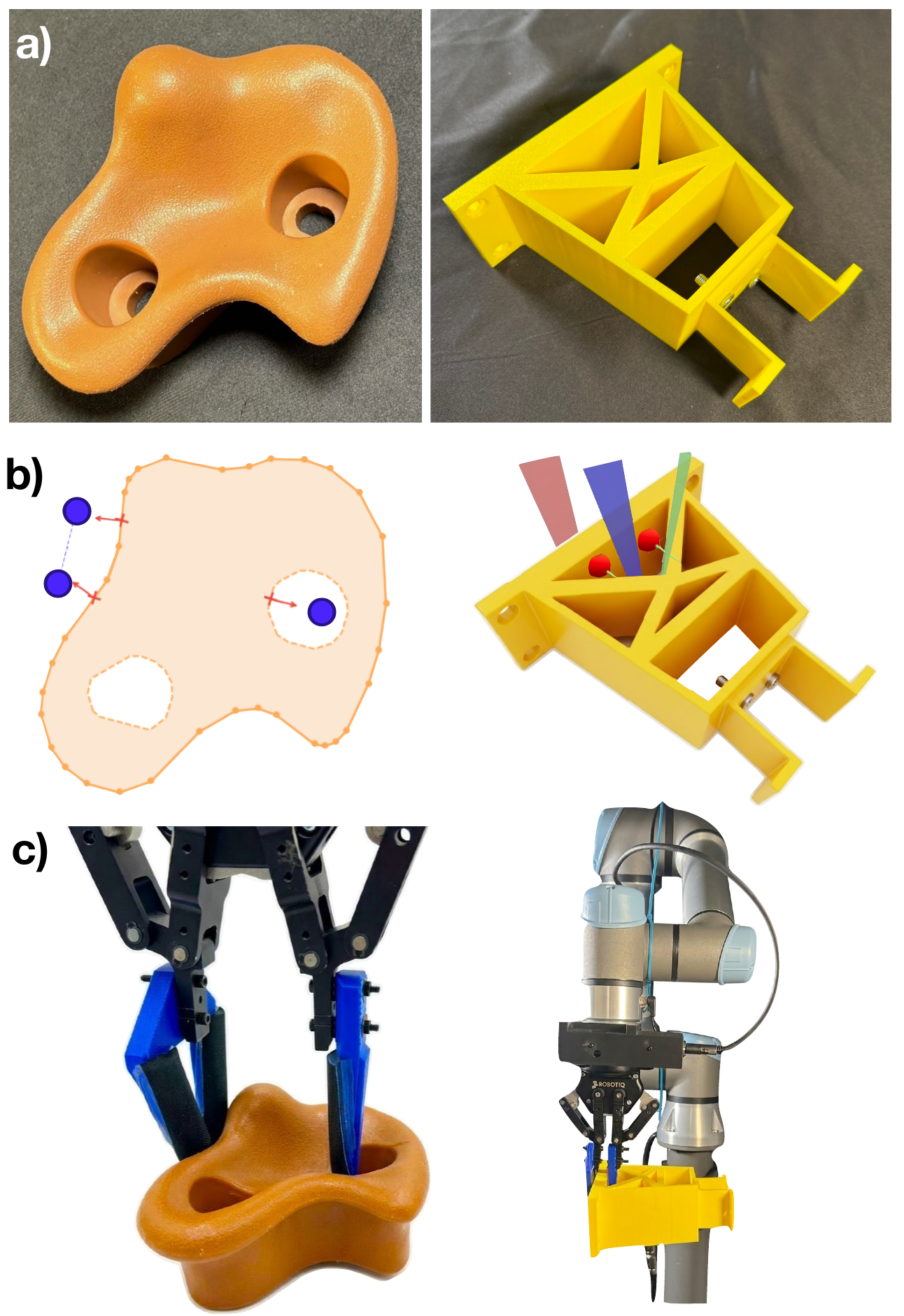}
    \caption{(a) Using a single top-down RGB-D observation of an object, (b) the planner receives a 2D segmentation mask and point cloud, samples grasps, and computes grasp wrench space metrics. 
    GET-2D-1.0 plans grasps on a 2D polygon footprint of the object and GET-3D-1.0 plans on a 3D mesh.
    (c) A UR5e robot arm equipped with the GET gripper executes the highest-rated grasp.}
    \label{fig:teaser}
\end{figure}

2) GET-3D-1.0: A mesh-based 3D grasp planner that efficiently samples candidate grasps and computes force closure and Ferrari-Canny grasp metrics.

3) Physical experiments comparing with a baseline planner that suggest GET-2D-1.0 outperforms a baseline 2D Bounding Box method by over 40\% on lift success, shake survival, and force resistance. Experiments also suggest that GET-3D-1.0 provides slight further improvements, but requires more computation time.
\begin{figure}
    \centering
    \includegraphics[width=1\linewidth]{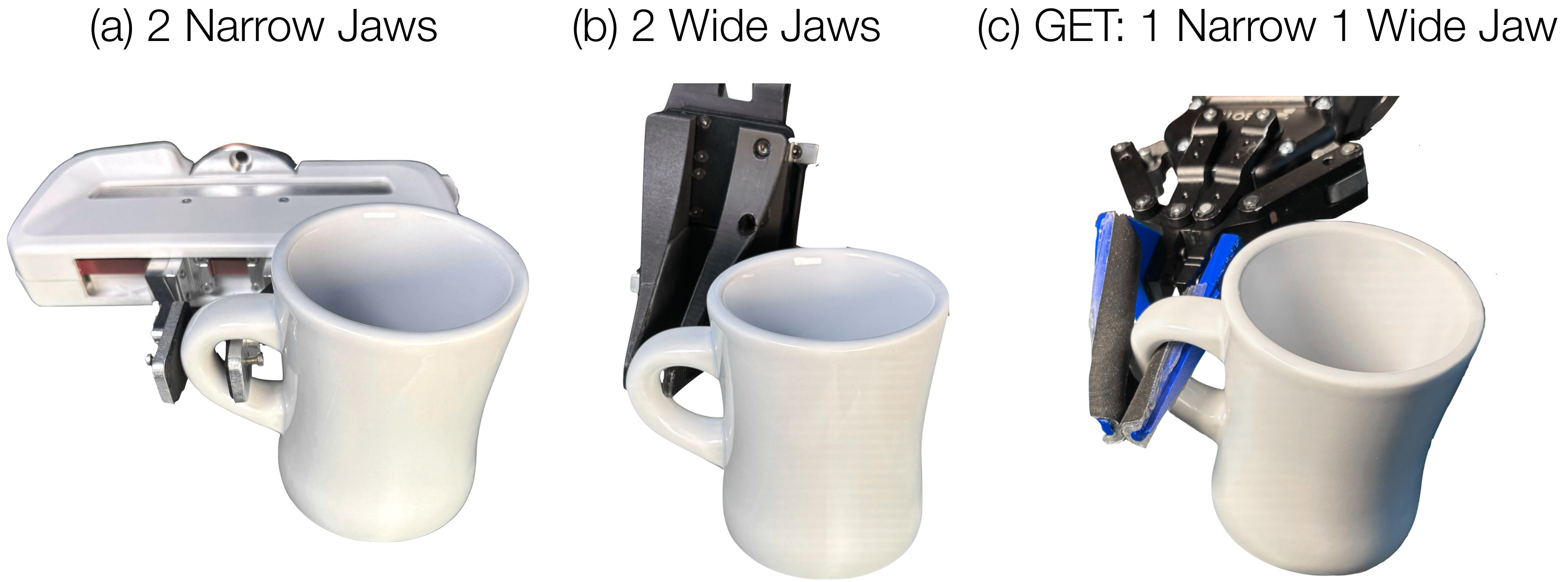}
    \caption{\textbf{3 Parallel Gripper Jaw Designs} (a) Narrow jaws are susceptible to torque about the grasp axis. (b) Wide jaws cannot access small concavities. (c) The GET Asymmetrical Gripper provides asymmetric geometry that addresses both limitations.}
    \label{fig:comparison}
\end{figure}
\section{RELATED WORK}

\subsection{Hardware Design}
As illustrated in Fig.~\ref{fig:comparison}, a traditional symmetric narrow parallel-jaw gripper offers mechanical simplicity, but restricts grasping to two antipodal contact points with poor torque resistance. 
A few notable hardware innovations have diverged from the traditional 2 narrow-jaw gripper. One popular setup is the YAM parallel-jaw gripper with 2 wide V-shaped jaws, but this design still faces the fundamental limitations of antipodal point contacts.

To overcome these limits, some researchers have explored hands with higher degrees of freedom~\cite{cut1985, sal1982}. Sunday Robotics introduced a 4-DoF 3 Finger gripper~\cite{sunday2025act1}.

The GET gripper, proposed in 2025 by Burgess and Adelson, is the primary inspiration and hardware backbone for this research~\cite{burgess2025get}. The researchers introduced a novel 1-DoF asymmetric design aimed at securely grasping objects of many shapes and sizes. Mounted on a standard parallel-jaw actuator, they design a two-against-one configuration, where a wide V-shaped jaw opposes a single narrow jaw. Burgess and Adelson primarily focus on the hardware design of the GET gripper, emphasizing its ability to conform to object geometries and form secure grasps better than traditional symmetric designs. 

\begin{figure}
\centering
\includegraphics[width=.95\linewidth]{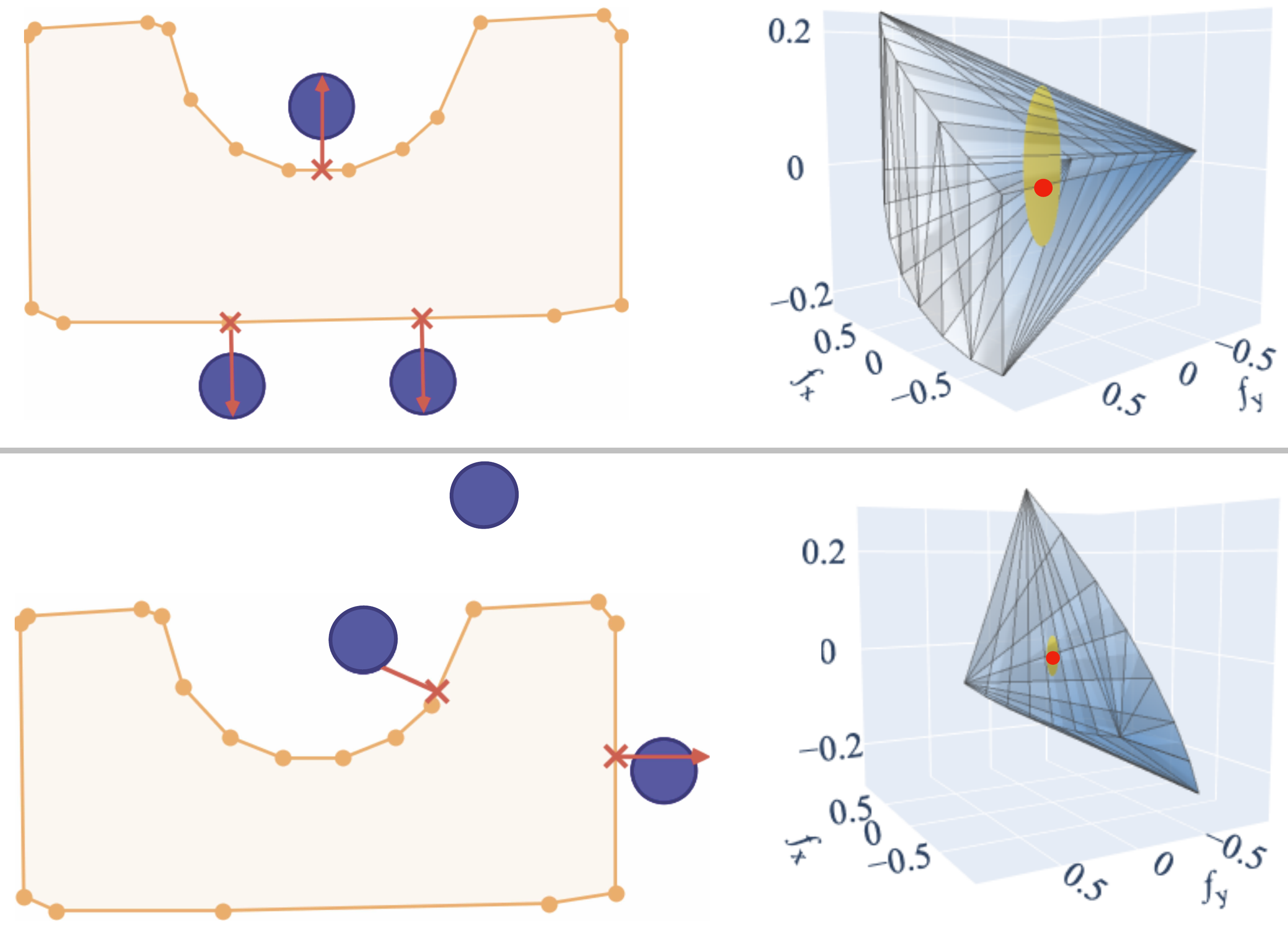}
\caption{Conceptual visualization of the Grasp Wrench Space (GWS) and the Ferrari-Canny metric ($\varepsilon$-distance) for a high-quality grasp (top) and a low-quality grasp (bottom). Gripper fingers appear as blue circles with each contact point marked by a red X. Torque normalized by object length is plotted along the vertical axis. The radius of the yellow ellipse is the Ferrari-Canny metric, and the red circle denotes the origin. A grasp is in force closure when the origin is contained in the interior of the GWS.}
\label{fig:ferrari_canny}
\end{figure}

\subsection{Analytical Grasp Planning}
Grasp planning relies on analytical mechanics, specifically the geometric analysis of contact points and their ability to resist external forces~\cite{joh1985, mur1994}. Salisbury and Roth (1982) pioneered the use of screw theory in robotics, establishing that a necessary and sufficient condition for force closure is that the primitive contact wrenches positively span the entire wrench space~\cite{sal1982}. Building upon this foundation, Nguyen (1988) developed algorithms for constructing force-closure grasps based on object geometry~\cite{ngu1988}. Furthering this, Ferrari and Canny (1992) introduced quantitative metrics to evaluate grasp quality based on the Grasp Wrench Space (GWS), specifically the minimum force required to resist arbitrary external perturbations~\cite{fer1992}. While these analytical methods provide strict mathematical guarantees, they often struggle in noisy, real-world environments where precise 3D models and friction coefficients are unknown~\cite{how1996, bic1995}. To address this uncertainty, frameworks like Dex-Net 1.0 have adapted these classical metrics by applying randomized perturbations to evaluate grasp robustness, an approach we build on by using force closure and the Ferrari-Canny metric~\cite{mah2017}.
    
\subsection{Data-Driven Grasp Planning}

Machine learning can predict grasp success from sensor data~\cite{kri2018}. Mahler et al. pioneered the Dexterity Network (Dex-Net) 2.0, 3.0, and 4.0 series, which used synthetic datasets generated by robust analytical metrics to train deep neural networks~\cite{mah2017, mah2018, mah2019}. Dex-Net 1.0 introduced a cloud-based network of 10,000 3D objects, utilizing a Multi-Armed Bandit model with rewards to leverage prior grasps and accelerate robust grasp planning~\cite{mah2017}. Dex-Net 2.0 advanced this by training a Grasp Quality Convolutional Neural Network (GQ-CNN) on 6.7 million synthetic point clouds and analytic metrics, enabling the rapid prediction of robust parallel-jaw grasps directly from depth images~\cite{mah2017}. Recognizing the need for diverse end-effectors, Dex-Net 3.0 expanded the framework to vacuum-based suction cup grippers, proposing a compliant suction contact model to evaluate seal formation and wrench resistance~\cite{mah2018}.

Parallel to Dex-Net, Fang et al. developed GraspNet-1Billion, a large-scale benchmark for general object grasping containing 97,280 RGB-D images and over one billion densely annotated 6-DoF grasp poses~\cite{fang2020}. This dataset provided the scale necessary to train highly accurate 3D grasp pose detection networks capable of operating in dense clutter~\cite{fang2020}. 

More recent works like ``Learning to grasp by playing with random toys"~\cite{niu2025learning}, show how demonstrations on a limited set of objects can be used to improve general grasping performance. 
This work focuses on the underlying analytical foundations that many data-driven methods rely on for generating training labels, adapting them to the asymmetric geometry of the GET gripper \cite{burgess2025get}.

\begin{figure*}
    \centering
    \includegraphics[width=1\linewidth]{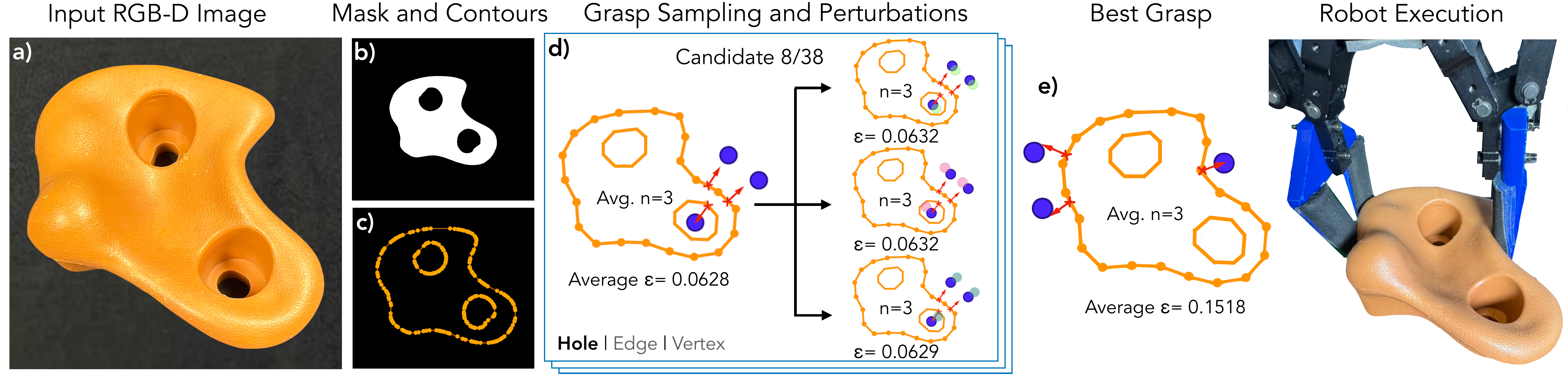}
    \caption{(a) The 2D grasp planner begins by capturing an overhead RGB-D image of the object. (b) The object is converted to a binary mask via color thresholding.  (c) The planner extracts the contours of the mask and uses the Teh-Chin and Ramer-Douglas-Peucker algorithms to transform the contours to a simplified polygon.  (d)
    Then, the planner generates candidate grasps and evaluates their force closure, number of contacts made ($n$), and Ferrari-Canny metrics ($\varepsilon$) across multiple positional perturbations, and averages the results to obtain the force closure rate, average number of contacts made, and average epsilon metric. 
    (e) The grasp scored highest by the metrics is chosen to be executed.}
    \label{fig:method}
\end{figure*}

\section{PROBLEM DEFINITION}

\textit{Definitions and Objective:} The grasping problem is characterized by a state $x = (\mathcal{O}, c_o, T_o, \gamma)$, where $\mathcal{O}$ represents the object mesh, $c_o \in \mathbb{R}^3$ is the center of mass, $T_o \in SE(3)$ denotes the 3D pose, and $\gamma \in \mathbb{R}$ is the friction coefficient~\cite{mah2017}. 
The 2D planner receives an RGB-D observation $y$ and outputs a top-down grasp $u = (p, \phi, z) \in \mathbb{R}^2 \times S^1 \times \mathbb{R}$, specifying the target gripper position $p$, grasp angle in the table plane $\phi$, and depth $z$. The 3D planner receives $(y, \mathcal{O})$ as input, and outputs a grasp $u \in SE(3)$. The objective for both planners is to maximize $Q(u, y)$, a measure of grasp robustness under uncertainty. The planners seek an optimal policy $\pi^*(y) = \text{argmax}_{u \in \mathcal{C}} Q(u, y)$ over the set of collision-free gripper configurations~\cite{mah2017}.

\textit{Assumptions:} We assume the target object is a rigid body resting in a singulated configuration on a planar surface of known height, and that the object can be reliably segmented from the background. Object-jaw contacts are modeled using a Point Contact with Friction (PCWF) model governed by Coulomb's law. We also assume a calibrated stereo camera with known extrinsic and intrinsic parameters.

A grasp is considered secure if it achieves force closure, which requires that the origin lies in the interior of the Grasp Wrench Space (GWS). The GWS is defined as the convex hull of all wrenches that can be generated at the contacts for a given contact model, subject to a unit norm constraint on normal forces ~\cite{fer1992}. When a grasp is in force closure, the contacts can collectively resist arbitrary external wrenches applied to the object.

 The Ferrari-Canny metric~\cite{fer1992}, or $\varepsilon$-distance, is defined as the radius of the largest hypersphere centered at the origin that can be fully inscribed within the GWS boundary. A larger $\varepsilon$ value indicates a more robust grasp capable of resisting greater wrenches. 

The GET gripper asymmetric 2-against-1 geometry generally forms three distinct contacts, increasing the volume of the GWS convex hull compared to traditional 2 narrow jaw grippers. (See Fig. \ref{fig:ferrari_canny} for geometric visualization of the GWS and Ferrari-Canny).


\section{GET GRIPPER MECHANICAL DESIGN}
\label{sec:mech-design}
Burgess and Adelson present a GET gripper configuration 3D printed with Markforged Onyx material and with round, custom-molded silicone gel pads and internal cameras for tactile sensing. We modify the design to use high-friction tape on three layers of Very High Bond tape, which enables slight finger deformation. This reduces fabrication cost and complexity, while maintaining the advantages of the GET gripper geometry. We also incorporate a reinforced backing to increase rigidity, revised mounting for compatibility with a Robotiq 2F-85 gripper, and fabrication by 3D printing with PLA. The gripper is scaled down to have 6.5 cm of usable finger length, similar to the YAM and ALOHA grippers~\cite{zhao2023aloha}. The maximum gripper width is 9.5 cm (see Fig. \ref{fig:teaser}).

\begin{figure*}
    \centering
    \includegraphics[width=1\linewidth]{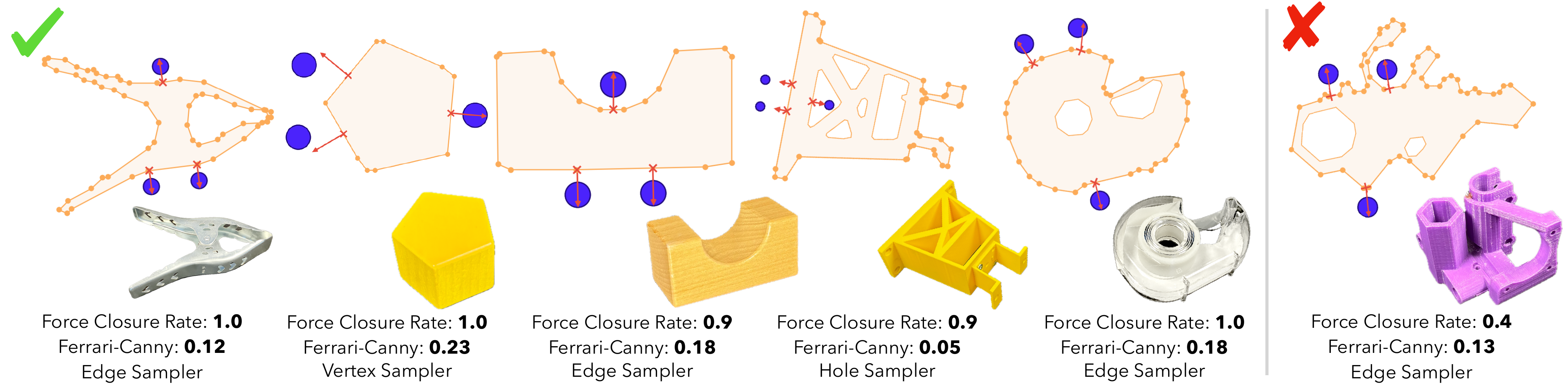}
    \caption{
    \textbf{GET-2D-1.0:}
    The 2D grasp planner models the GET gripper using three equally sized circles shown in blue. From their resulting contact points and contact normals shown in red, the planner computes grasp wrench space metrics. The planner samples perturbed positions from the original candidate grasp and computes the percentage of those perturbed grasps in force closure and the average Ferrari-Canny metric across those positions. The grasp on the right is an example where the original candidate has a high Ferrari-Canny metric (0.13), but because the contacts are very close to sharp vertices, the metrics averaged over perturbed grasps are much lower. This causes a different grasp to be selected. 
    }
    \label{fig:examples_2d}
\end{figure*}

\section{GET-2D-1.0 Planner}

For rapid grasp execution, we first present GET-2D-1.0. The pipeline begins by capturing a top-down RGB-D image of the object using a single wrist-mounted camera. The planner creates a 2D segmentation mask of the object using color thresholding and retrieves contours using the Suzuki-Abe algorithm~\cite{suzuki1985}. A simplified polygon is created using the Teh-Chin algorithm~\cite{teh2002detection} to maintain important points, followed by Ramer-Douglas-Peucker ~\cite{ramer1972, douglas1973} to remove near-colinear vertices. The resulting polygon represents the top-down footprint of the object geometry and typically has between six and eighty vertices, depending on object complexity. To convert from this image space polygon to metric units, the planner estimates a depth-dependent pixels-per-meter scaling factor to account for perspective projection. A reference pixels-per-meter is calibrated at the table plane and is multiplied by $z_{obj}/z_{table}$ where $z_{obj}$ is the 20th percentile distance to the object obtained from the depth image and $z_{table}$ is the distance between the camera and table.

As shown in Fig. \ref{fig:examples_2d}, the GET gripper is modeled using three equally sized circles -- two for the wide jaw and one for the narrow jaw. Grasp candidates are rejected if: their initial pose results in those circles overlapping the object polygon, or if the necessary gripper opening exceeds the gripper maximum opening, or if there is not at least one contact on each jaw.
The planner uses three complementary sampling strategies (see Fig. \ref{fig:examples_2d}) to get a set of candidate grasps. The edge sampler places the wide jaw normal to each polygon edge at evenly spaced positions along the edge length. The vertex sampler adds grasp candidates straddling each vertex with the wide jaw and considers small angle differences from the bisector of the adjacent edge normals. The hole sampler places the narrow jaw into each hole with the gripper closing axis normal to each hole edge. 

Each grasp candidate is evaluated by sampling a fixed number of 2D translational perturbations from a Gaussian distribution and averaging grasp metrics computed at each perturbed position. The center of mass of the object is computed based on the 2D mask under a uniform density assumption. The grasps are sorted by their force closure rate, with the average number of contacts, then the average Ferrari-Canny metric used to tiebreak. The grasp that ranks highest is selected for execution.

\begin{figure}
    \centering
    \includegraphics[width=1\linewidth]{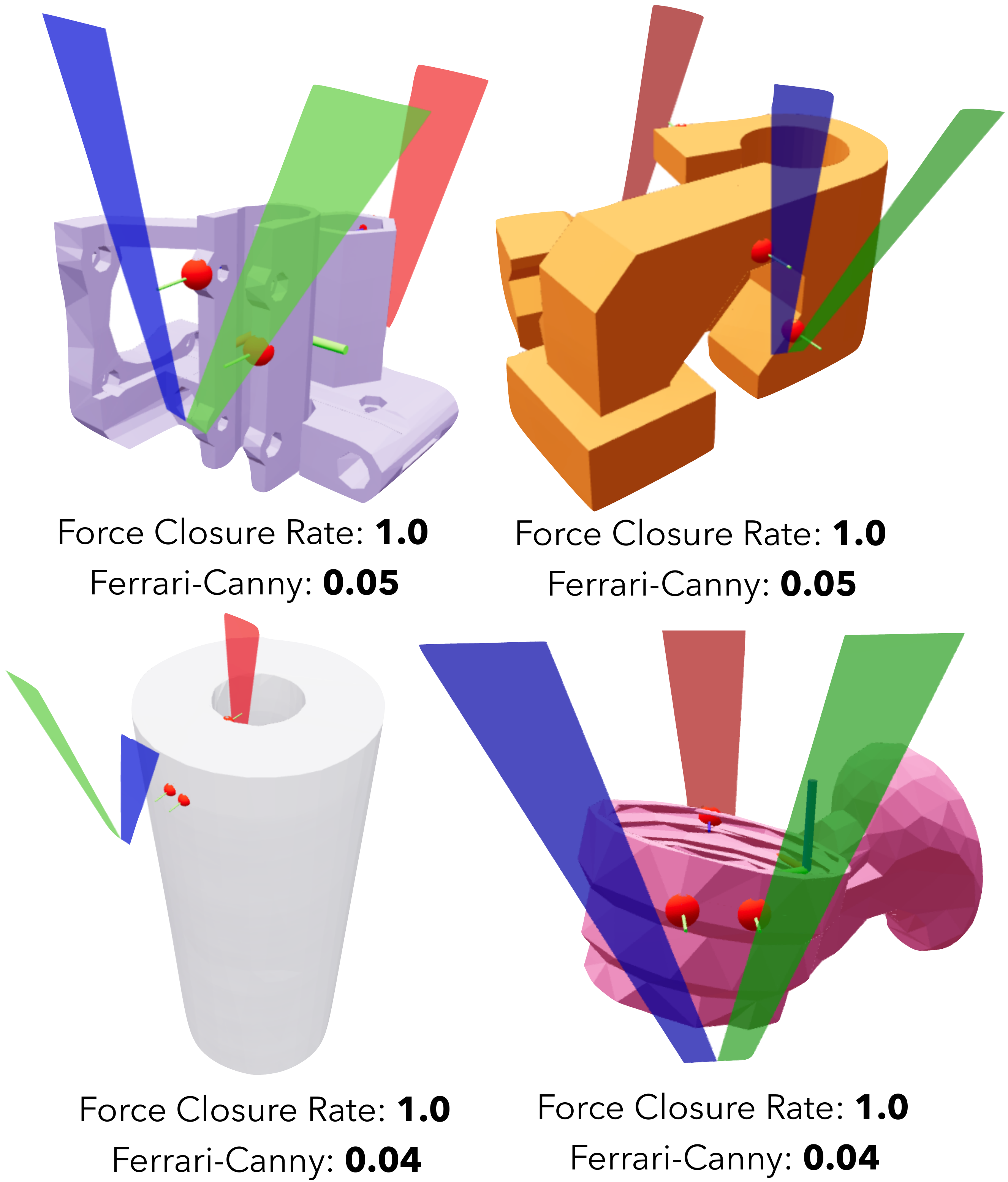}
    \caption{\textbf{GET-3D-1.0}: The planner samples a gripper pose parallel to an object face and casts rays from the contact surfaces to determine where the gripper will make contact with the mesh. The resulting contact points shown as red spheres are used to compute force closure and the Ferrari-Canny metric.} 
    \label{fig:examples_3d}
\end{figure}

\section{GET-3D-1.0 Planner}
We also implement GET-3D-1.0. The 3D perception pipeline captures a top-down point cloud from the depth camera and uses color thresholding to segment the object in the point cloud. The planner aligns a pre-acquired mesh in a matching stable pose to the object point cloud using the Iterative Closest Point algorithm~\cite{besl1992}.

The GET-3D-1.0 planner models the gripper as three trapezoidal contact surfaces as seen in Fig. \ref{fig:examples_3d}. To obtain a candidate grasp, the planner first randomly samples a mesh face weighted by face area and within a specified angular tolerance from vertical. The wide jaw of the gripper is oriented normal to the sampled face and centered at a random point sampled on the face. The gripper is placed with a uniformly random rotation about its closing axis and a uniformly random z-offset between zero and the table distance. A set of rays in a uniform grid is cast outward from the contact surfaces of the gripper along the gripper closing axis direction into the environment. For each ray, the planner computes the first intersection point with the mesh and intersection distance to the mesh.

The contact points on the object can be determined by finding the intersection point corresponding to the minimum ray-surface distance from each finger. If the distances for the two contact points on the wide jaw are imbalanced, only the closer contact point will be realized. The sampler minimizes this imbalance -- the absolute difference between the distance to each finger's contact point -- by optimizing the gripper rotation about its vertical axis. Specifically, the optimizer uses the derivative of the imbalance with respect to the gripper rotation to perform gradient descent with an adaptive learning rate until the imbalance is less than a small tolerance. This ensures the sampler finds grasps that will form three contact points. If the optimization fails to converge, the gripper is in collision with the mesh, or the gripper is too close to or below the table, the sample is rejected, and the sampling process restarts until a suitable three-contact configuration is found.

Each contact is modeled as a 3D point contact with friction, and its associated normal is the mesh normal at the contact point. The object center of mass is based on the 3D mesh under a uniform-density assumption. Grasp candidate quality is evaluated first using the force closure rate from samples with positional perturbations sampled from a Gaussian distribution. Then, for grasps that tie for the maximum force closure rate, the Ferrari-Canny metric is computed, and the grasp that ranks highest is the grasp that is executed. Computing the Ferrari-Canny metric only for the top grasps significantly reduces the planning time. Similarly, the planner will not compute perturbation metrics for a grasp candidate that is not in force closure and will stop computing perturbations for a grasp that falls below a force closure rate threshold. This again enables faster grasp sampling without affecting the final selected grasp under the chosen ranking procedure.

 \section{EXPERIMENTAL METHODOLOGY AND RESULTS}
To evaluate the capabilities of the GET-2D-1.0 and GET-3D-1.0 planners, we assess the quality of planned grasps in physical experiments using a robot shake test and a digital force gauge. We select a set of 10 objects and run two trials for each in different stable poses. The 10 objects are: clamp, arch, yellow camera mount, tape roll, purple mount (Fig. \ref{fig:examples_2d}), rock climbing hold (Fig. \ref{fig:method}), pink turbine housing, orange bar clamp, paper towel roll (Fig. \ref{fig:examples_3d}), and gearbox. Four of these objects come from the Dex-Net set of adversarial objects intended to be difficult to grasp with parallel jaw grippers.

Experiments are run with a UR5e 6-DoF robot arm equipped with a 2F-85 Robotiq gripper with the GET gripper jaws (as described in Section \ref{sec:mech-design}) attached. The gripper is set to its minimum force setting so that if an object slips, the grasp is not tightened. A Zed Mini stereo depth camera is attached to the wrist of the robot, and objects are placed on a flat table with a black tablecloth.

\subsection{Implementation Details}
The bounding box baseline (2D BBox) is implemented by finding the oriented bounding box of the segmentation mask and executing a grasp with the gripper closing direction perpendicular to the longer side and through the bounding box center. The grasp is executed 5 cm below the 80th percentile object height determined from the point cloud. This baseline is performed with the GET gripper as described in section \ref{sec:mech-design} and with two narrow and two wide flat jaws with friction tape applied.

For GET-2D-1.0, each contact circle is 1 cm in diameter. The spacing between the two fingers on the wide jaw is 3 cm, and the gripper has a max opening of 9.5 cm. The grasp sampler uses a max tolerance of 4 mm to model if both wide jaw circles make contact. For each candidate grasp, the planner averages metrics over 10 samples and perturbs the grasp coordinates by adding Gaussian noise with a standard deviation of 2 mm. The grasp height is determined the same way as in the bounding box baseline.

The GET-3D-1.0 planner assumes a pre-specified mesh in a stable pose matching the target object. For each stable pose, the planner samples 500 candidate grasps. GET-3D-1.0 optimizes the wide jaw contact balance for up to 30 iterations or until below the max imbalance tolerance of 1 mm. Positional perturbations use a standard deviation of 7 mm. The maximum grasp degree from vertical is 20 degrees, and the maximum rotation about the gripper closing axis is 30 degrees. 

Both planners model the friction coefficient as $\gamma=.75$.

\subsection{Shake Test and Force Gauge}
\begin{figure}
    \centering
    \includegraphics[width=1\linewidth]{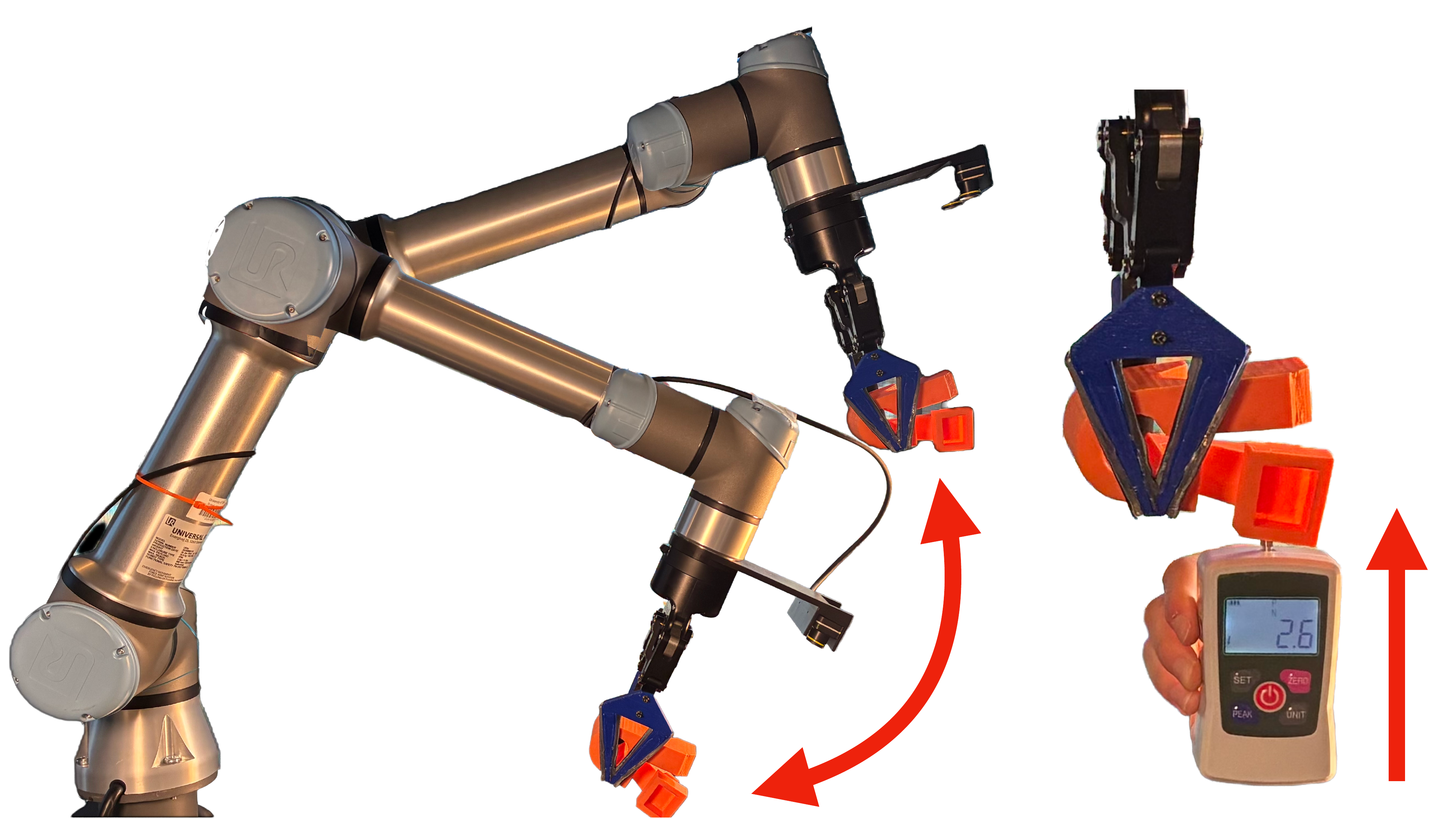}
    \caption{\textbf{Experimental setup for evaluating grasp quality. }(Left) Robot arm executes a dynamic shake test at $4.5\pi$ rad/s to verify stability. (Right) For grasps that survive the shake test, a digital force gauge measures resistance to an external force.}
    \label{fig:shake_test}
\end{figure}
 
After a successful grasp, defined by the object being lifted and held off the table, the robot performs an automatic shake test (Fig. \ref{fig:shake_test}) which enables a consistent evaluation across trials. The shake test is performed by moving the arm back and forth two times in a range of $0.6$ rad with a joint acceleration of $1.5 \pi$ rad/s, then $4.5\pi$ rad/s in the third joint (counting from the base joint). This test is then repeated after turning the gripper 90 degrees so that the force is exerted in a perpendicular direction.

For objects which remain in the gripper after the shake test, we use a digital force gauge~\cite{hojila2021gauge} (Fig. \ref{fig:shake_test}) and measure the peak force required to rotate the object 20 degrees about the gripper closing axis while pressing at the point on the object furthest from that axis. This enables finer grain evaluation of strong grasps. Objects which are dropped or which can be rotated with a force below the force gauge's minimal reading of 2 N are assigned a force of 0 N. 

\begin{table}
\centering
\caption{Physical Grasp Experiments}
\label{tab:success_metrics}
\setlength{\tabcolsep}{4pt}  
\begin{tabular}{lrrr}
\textbf{Method} & \textbf{Lift \%} & \textbf{Shake \%} & \textbf{Slip Force (N)} \\
\hline
2D BBox (Two Narrow) & 70\% & 55\% & 1.20 \\
2D BBox (Two Wide) & 50\% & 40\% & 1.46 \\
2D BBox (GET) & 65\% & 60\% & 2.50 \\
GET-2D-1.0 & \textbf{95\%} & \textbf{90\%} & \textbf{3.95} \\
Improvement over 2D BBox & 46\% & 50\% & 58\% \\
\hline \\[-2mm]  
GET-3D-1.0 & \textbf{100\%} & \textbf{95\%} & 3.20 \\
Improvement over GET-2D-1.0 & 5\% & 6\% & -19\% \\
\hline
\end{tabular}
\end{table}

\subsection{Quantitative Results}
As shown in Table \ref{tab:success_metrics}, we evaluate each grasp planner on three metrics: lift success, shake success, and average force. Lift success is defined as the percentage of grasps that result in the object being successfully lifted and held off the table. Shake success is the percentage of trials where the object remains in the gripper after the shake test is executed. Average force is the average peak force gauge reading across the 20 trials. In all three metrics, the 2D grasp planner outperforms the bounding box baseline by more than 40\%.

The bounding box method is unable to take advantage of holes in objects like the yellow mount in Fig. \ref{fig:teaser} or paper towel roll in Fig. \ref{fig:examples_3d}. This leads to grasps that exceed the maximum gripper opening, resulting in lift failure.
The bounding box planner also often generates grasps that fail to engage all three contact surfaces of the GET gripper, leading to objects slipping.

The hole sampler in the 2D grasp planner allows it to sample and evaluate grasps that insert the narrow jaw into holes in the polygon, avoiding a failure case of the bounding box planner. 
In the trial where the 2D grasp planner failed to lift an object, the grasp was executed above the target surface on the object due to the use of a fixed height offset from the object.

The 2D Bounding Box method runs in under a millisecond, while GET-2D-1.0 runs in an average of 683 ms, with planning time varying based on the number of vertices on the object polygon.

GET-3D-1.0 successfully lifted all objects, with all but one object passing the shake test. However, GET-3D-1.0 requires an object mesh and is substantially slower with an average planning time of 17 seconds because the planner samples a larger number of grasps and computes contact points in 3D. This justifies using GET-2D-1.0 in practice as it is fast and comparable to the more complex 3D planner, even outperforming it in average slip force.

\section{LIMITATIONS AND FUTURE WORK}

This work presents two novel grasp planning algorithms for the GET asymmetric gripper. While experiments suggest these methods are effective, limitations remain. Physical experiments are limited to non-black objects in singulated configurations on a black surface.  GET-3D-1.0 requires a mesh as input and is computationally expensive. In future work, we will explore generating datasets for learned grasp planners.


\end{document}